\documentclass[10pt,twocolumn,letterpaper]{article}

\usepackage[pagenumbers]{cvpr} 

\definecolor{cvprblue}{rgb}{0.21,0.49,0.74}
\usepackage[pagebackref,breaklinks,colorlinks,allcolors=cvprblue]{hyperref}

\usepackage{multirow}
\usepackage{algorithm}
\usepackage{algpseudocode}
\usepackage{graphicx}
\usepackage{bbding}

\def\confName{CVPR}
\def\confYear{2025}

\title{PackDiT: Joint Human Motion and Text Generation via Mutual Prompting}

\author{Zhongyu Jiang \quad
Wenhao Chai \quad
Zhuoran Zhou \quad
Cheng-Yen Yang \quad \\
Hsiang-Wei Huang \quad
Jenq-Neng Hwang \vspace{4pt}\\  University of Washington \vspace{4pt} \\
{\tt\small \{zyjiang, wchai, zhouz47, cycyang, hwhuang , hwang\} @ uw.edu}
}

\begin{document}


\twocolumn[{
    \maketitle
    \vspace{-3em}
    \renewcommand\twocolumn[1][]{#1}
    \begin{center}
    \centering
    \includegraphics[width=0.93\textwidth]{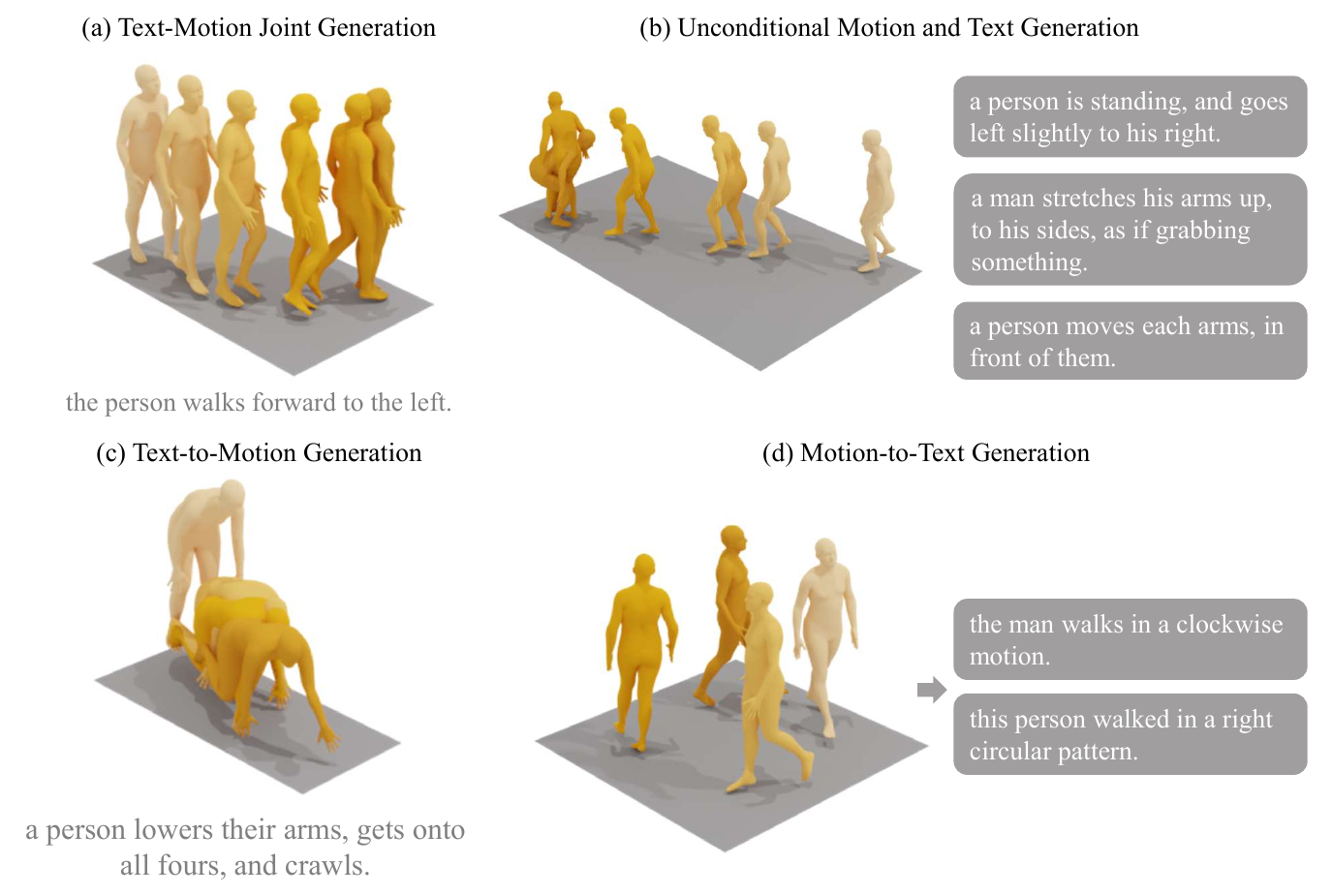}
    \captionof{figure}{PackDiT is able to conduct four different motion-related generation tasks, \eg, Text-Motion Joint Generation, Unconditional Motion and Text Generation, Text-to-Motion Generation, and Motion-to-Text Generation. All data are generated with PackDiT.}
    \label{fig:teaser}
\end{center}
\vspace{1em}
}]

\begin{abstract}

Human motion generation has advanced markedly with the advent of diffusion models. Most recent studies have concentrated on generating motion sequences based on text prompts, commonly referred to as text-to-motion generation. However, the bidirectional generation of motion and text, enabling tasks such as motion-to-text alongside text-to-motion, has been largely unexplored. This capability is essential for aligning diverse modalities and supports unconditional generation. In this paper, we introduce \textbf{PackDiT}, the first diffusion-based generative model capable of performing various tasks simultaneously, including motion generation, motion prediction, text generation, text-to-motion, motion-to-text, and joint motion-text generation. Our core innovation leverages mutual blocks to integrate multiple diffusion transformers (DiTs) across different modalities seamlessly.
We train PackDiT on the HumanML3D dataset, achieving state-of-the-art text-to-motion performance with an FID score of $0.106$, along with superior results in motion prediction and in-between tasks. Our experiments further demonstrate that diffusion models are effective for motion-to-text generation, achieving performance comparable to that of autoregressive models.

\end{abstract}    
\begin{table*}[t]
\caption{Comparison of recent state-of-the-art methods on diverse motion-relevant tasks. \textit{Random Motion} and \textit{Random Text} represent unconditional generation of motions and motion descriptions. \textit{Joint Gen} means the joint generation of motion and motion descriptions.}
\resizebox{0.98\textwidth}{!}{%
\begin{tabular}{lccccccccccc}
\toprule
Methods & Text-to-Motion & Motion-to-Text & Motion Pred. & Motion In-Between & Random Motion & Random Text & Joint Gen.
\\ \midrule
T2M-GPT~\cite{zhang2023t2mgpt} & \checkmark & - & - & - & \checkmark & - & - \\
MLD~\cite{chen2023mld} & \checkmark & - & - & -& \checkmark & - & - 
  \\
TM2T~\cite{guo2022tm2t}  & \checkmark & \checkmark  & - & - & - & - & - 
  \\
MDM~\cite{tevet2023mdm} & \checkmark   & -  & \checkmark & \checkmark & \checkmark & - & - \\
MotionDiffuse~\cite{zhang2024motiondiffuse} & \checkmark   & -  & \checkmark & \checkmark & \checkmark & - & - 
  \\
LMM~\cite{zhang2024large} & \checkmark& - & \checkmark& \checkmark& \checkmark& - & - \\
MotionGPT~\cite{jiang2024motiongpt} & \checkmark& \checkmark& \checkmark& \checkmark& \checkmark& \checkmark & - \\
\midrule
PackDiT~(ours) & \checkmark& \checkmark& \checkmark& \checkmark& \checkmark & \checkmark & \checkmark
\\\bottomrule
\end{tabular}%
}
\label{tab:task}
\end{table*}

\section{Introduction}


Human motion capture is widely used across multiple industries, including film production, video game development, and virtual reality (VR). However, setting up a motion capture studio is expensive, and the quality of the captured motion heavily depends on the actors' performance. With advancements in diffusion models~\cite{ho2020ddpm, song2020ddim, song2019score, song2020score}, recent years have seen substantial progress in Motion Generation~\cite{zhu2023human, tevet2023mdm, guo2022tm2t, zhang2024motiondiffuse, jiang2024motiongpt, zhang2024large, zhang2023t2mgpt, tseng2023edge}, which aims to automatically generate rich, realistic human motion sequences.

Motion generation encompasses the production of human motion sequences with or without conditions from other modalities, such as action classes, text, audio, music, and speech. Among these, text stands out for its ability to convey detailed information about actions, speeds, directions, and goals, either explicitly or implicitly. For instance, HumanML3D~\cite{guo2022humanml3d} is a comprehensive text-to-motion generation dataset that provides well-annotated text-motion pairs derived from HumanAct12~\cite{guo2020humanact12} and AMASS~\cite{mahmood2019amass}. Recent studies leverage such datasets to explore diffusion-based models for text-to-motion generation and autoregressive models for motion-to-text understanding.

However, few methods can do both text-to-motion and motion-to-text generation. Recently, MotionGPT~\cite{jiang2024motiongpt} uses the auto-regressive paradigm to achieve this goal. Addressing the challenges of effectively generating and integrating motion and text, we propose a novel framework, \textbf{PackDiT}, the first diffusion-based text-motion joint generation model. PackDiT stands out for its flexibility and capability to handle multiple tasks within a unified architecture, leveraging two independent Diffusion Transformers (DiTs), Motion DiT and Text DiT, with mutual blocks and multi-stage training strategies. PackDiT is initially pre-trained unconditionally, then jointly trained and fine-tuned, enhancing fidelity and alignment.


We evaluate PackDiT on the HumanML3D dataset~\cite{guo2022humanml3d} across a range of tasks and corresponding metrics. Compared to other state-of-the-art text-to-motion generative models, PackDiT achieves superior performance on the FID metric with fewer parameters. Additionally, PackDiT demonstrates leading performance in motion prediction and in-between tasks. Notably, we are the first to show that a diffusion-based generative model can perform motion-to-text generation, achieving comparable results to large language models (LLMs) trained on extensive text corpora.

In particular, we make the following contributions:
\begin{itemize}[leftmargin=7.5mm]
\setlength{\itemsep}{2pt}
\item We are the first diffusion-based model that can accomplish diverse motion-relevant tasks, including text-to-motion, motion-to-text, and motion-text joint generation, \etc.
\item We show that by adding mutual blocks between text and motion diffusion generative models (\eg DiT), we can easily package two separated models to achieve good joint generation ability.
\item Our experiments show that our proposed method achieves state-of-the-art text-to-motion performance with FID as $0.106$, as well as the motion prediction and motion in-between tasks.
\end{itemize}

\section{Related Works}

\subsection{Diffusion Model}
For many years, researchers have been eager to find an effective method to generate various kinds of data, \eg, text, images, audio, and \etc~ After the creations of VAE~\cite{kingma2013vae}, GAN~\cite{goodfellow2014gan}, Normalizing Flows~\cite{rezende2015normalflow} and \etc, diffusion models~\cite{ho2020ddpm, song2020ddim, song2019score, song2020score,chai2023stablevideo,bao2023unidiffuser,cao2023difffashion,cao2023image} are proposed and shown to provide the best quality of generated results by training the model to gradually denoise the randomly initialized noise data and generate the final result. Based on diffusion probabilistic model (DPM)~\cite{sohl2015dpm}, Ho \etal proposed denoising DPM (DDPM), which utilizes U-Net~\cite{ronneberger2015unet} to denoise the noisy data step-by-step to recover the original data.  During forward diffusion of a DDPM, noisy data are generated by adding Gaussian noise to the original data step-by-step. In contrast, reverse diffusion aims to predict and remove the added Gaussian noise and gradually recover the original data.  DDIM~\cite{song2020ddim} is then proposed to accelerate the reverse diffusion of DDPM by skipping certain steps, and Score Matching Network~\cite{song2020score} takes advantage of Stochastic Differential Equations (SDE) to build a more general and effective diffusion pipeline. To further scale up the diffusion model, Peebles \etal\cite{peebles2023dit} propose the Diffusion Transformer (DiT), which utilizes transformers as the backbone of diffusion models.

However, all previous works only focus on single-modality generation, while joint multi-modal generation is also critical for synthetic data generation and real applications. UniDiffuser~\cite{bao2023unidiffuser} proposes a unified transformer-based diffusion model for joint Image-Text generation, which integrates text and image tokens into a unified diffusion model. Ruan \etal\cite{ruan2023mm} propose a joint multi-modal generation pipeline for video-audio generation via feature fusing and rotation layers.
Compared to other joint generation pipelines~\cite{suzuki2016joint, bao2023unidiffuser, ruan2023mm, tang2024any2any}, our approach does not necessitate training a unified generation model for all modalities. Instead, we employ independent generation models for each modality, integrating them through cross-attention layers. This design enhances the flexibility of PackDiT, handling joint generation tasks more effectively and efficiently.

\begin{figure*}[t]
    \centering
    \includegraphics[width=0.93\linewidth]{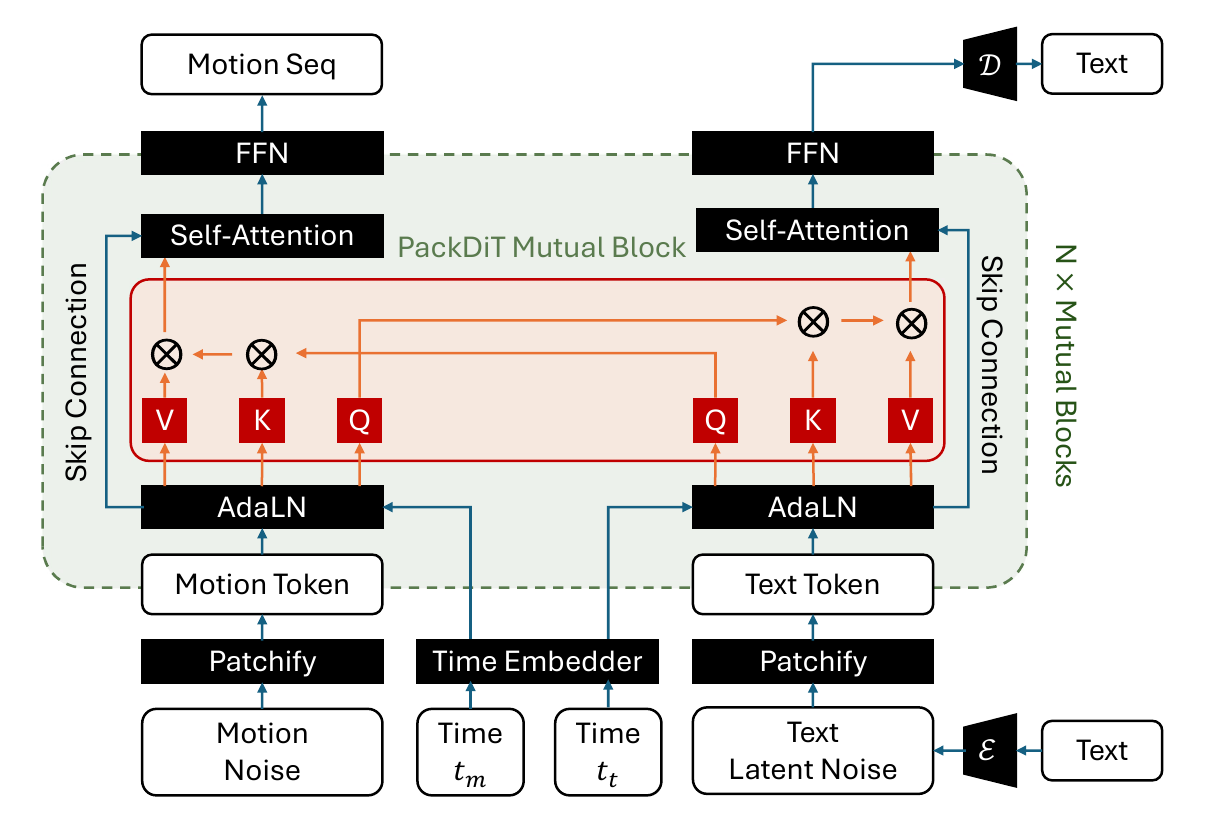}
    \caption{The architecture of PackDiT, where there are two independent DiTs for Motion and Text generation. By enabling and disabling the cross-attention layers in-between, PackDiT can solve almost all motion and text-related generation tasks, including text-to-motion, motion-to-text, motion prediction, motion in-between, random motion and text generation, and joint motion-text generation.}
    \label{fig:pipeline}
\end{figure*}

\subsection{Text-condition Human Motion Generation}
Human motion generation aims to generate realistic and controllable human pose sequences. Usually, people adopt SMPL~\cite{SMPL2015} or keypoint~\cite{chai2023global,jiang2024back,liu2023posynda} as the representation of 3D human pose instead of 3D keypoint joints. Researchers have been exploring using text, action, audio, music, or even scenes and objects as the conditions to guide the human motion generation. Among all those conditions, text has a remarkable capacity to convey information related to various actions, speeds, directions, and destinations, either explicitly or implicitly. This feature makes the text an appealing medium for generating human motion. Text2Action~\cite{ahn2018text2action} is the first to leverage GAN to generate a variety of motions from a given natural language description. Recently, diffusion models have been adopted to motion generation tasks successfully as well~\cite{dabral2023mofusion, guo2024generative, hoang2024motionmix,zhang2024motiondiffuse,zhang2024large}. However, although these methods have achieved excellent results in motion generation, they cannot simultaneously accomplish the task of action understanding, such as motion-to-text. Recently, MotionGPT~\cite{jiang2024motiongpt} uses the autoregressive paradigm of transformers to unify text-to-motion and motion-to-text within a single framework. In this paper, we use diffusion for the same purpose for the first time. As shown in Tab~\ref{tab:task}, compared to MotionGPT, our proposed framework additionally accomplishes joint generation.

\section{Method}
\label{sec:method}
To simulatneously tackle the motion-to-text, text-to-motion, and joint generation issues, we propose PackDiT, which is a flexible motion and text generation pipeline and consists of two DiT models, i.e., Motion DiT, $D_\mathcal{M}$ and Text DiT, $D_\mathcal{T}$. As shown in Table~\ref{tab:task}, compared with previous works based on diffusion models or large language models, PackDiT is able to achieve the most motion- and text-related generation tasks. For motion representations, we follow \cite{guo2022humanml3d, jiang2024motiongpt} and represent motion of frame $i$ as $\mathcal{M}_i \in \mathbb{R}^{263}$, where $\mathcal{M}_i = \{R_i, h^r_i, v^r_i, J_r, J_p, J_v, F\}$. $R_i$ is the global rotation of the human body, $h^r_i$ is the height of the root point, $v^r_i$ is the velocity of the root point, $J_r$ is the relative rotation of each of joints, $J_p$ is the position of each joint in canonical view, $J_v$ is the velocity of each joint, and $F$ represent each foot is on the ground or now. Following UniDiffuser~\cite{bao2023unidiffuser}, a text encoder, BERT~\cite{devlin2018bert}, and a text decoder, GPT-2~\cite{radford2019gpt2}, are utilized for text generation. 

\subsection{Unconditional Motion Generation}

To build a flexible multi-task joint generation diffusion pipeline, all inputs and outputs should be formulated into tokens, which allow us able to integrate mutual blocks for information exchange between Text and Motion. Therefore, we adopt DiT~\cite{peebles2023dit} as our baseline diffusion model. As shown on the left of  Fig~\ref{fig:pipeline}, for each timestep, $t_i$, motion representations $\mathcal{M} = \{\mathcal{M}_0, \dots, \mathcal{M}_n\}$ are converted to motion tokens after Patchify. Then, the Motion DiT, $D_\mathcal{M}$, generates the noise. The training procedure follows DDIM~\cite{song2020ddim}. Therefore, the training loss of $D_\mathcal{M}$ is

\begin{equation}
    \mathcal{L}_{\mathcal{M}} = \|\epsilon - D_\mathcal{M}(\sqrt{\overline{\alpha}_{t}}\mathcal{M} + \sqrt{1 - \overline{\alpha}_{t}}\epsilon, t)\|^2_2.
\end{equation}

\subsection{Unconditional Text Generation}

Following UniDiffuser~\cite{bao2023unidiffuser}, to make the diffusion model generate texts, we utilize a text encoder, $T_{enc}$, and a text decoder, $T_{dec}$, for better text generation quality. As shown on the right of Fig~\ref{fig:pipeline}, similar to our motion diffusion model, after encoder and forward diffusion, the text latent noise is fed to Text DiT, $D_\mathcal{T}$. After the reverse diffusion, all generated text tokens are treated as the prefix of the text decoder and used to generate the corresponding texts. Following DDIM~\cite{song2020ddim}, the training objective of $D_\mathcal{T}$ is 

\begin{equation}
    \mathcal{L}_{\mathcal{T}} = \|\epsilon - D_\mathcal{T}(\sqrt{\overline{\alpha}_{t}}\mathcal{T} + \sqrt{1 - \overline{\alpha}_{t}}\epsilon, t)\|^2_2.
\end{equation}

However, the diffusion model struggles to directly generate text tokens because they are discrete and have relatively high dimensions. To reduce the dimensionality of the encoded text tokens from the text encoder and create more continuous text representations, we further train an additional projection model, $P$, which projects the encoded text tokens to the text latent tokens, $\mathcal{T}$, used in $D_\mathcal{T}$, and re-projects the generated text tokens to the text decoder.


\subsection{Mutual Prompting}

Compared with other joint generation pipelines~\cite{suzuki2016joint, bao2023unidiffuser, ruan2023mm, tang2024any2any}, our pipeline does not require training a unified generation model for all modalities. Instead, the generation models for each modality operate independently and are integrated using mutual blocks. This architectural choice makes PackDiT significantly more flexible and capable of handling joint generation tasks more efficiently. Each modality-specific model can be optimized independently, allowing for specialized fine-tuning and improvements without affecting the entire system. This flexibility extends to scaling the system for new modalities, as new generation models can be added without extensive reconfiguration of the existing pipeline. As a result, PackDiT offers a robust and adaptable diverse joint generation.

During training or inference, the intermediate tokens, $M$ and $T$, from $D_\mathcal{M}$ and $D_\mathcal{T}$ generates $q_\mathcal{M}$, $k_\mathcal{M}$, $v_\mathcal{M}$, $q_\mathcal{T}$, $k_\mathcal{T}$, and $v_\mathcal{T}$. Then, the mutual blocks are operated:

\begin{align}
    \mathcal{M} &= \mathcal{M} + \text{Softmax}(\frac{q_\mathcal{M}k_\mathcal{T}^\top}{\sqrt{d_k}})v_\mathcal{T}, \\
    \mathcal{T} &= \mathcal{T} + \text{Softmax}(\frac{q_\mathcal{T}k_\mathcal{M}^\top}{\sqrt{d_k}})v_\mathcal{M}.
\end{align}

The cross-attention layer is inserted into each self-attention-based DiT block to fuse the information from all modalities and achieve flexible training or inference.

\begin{figure*}[t]
\centering
\includegraphics[width=\textwidth]{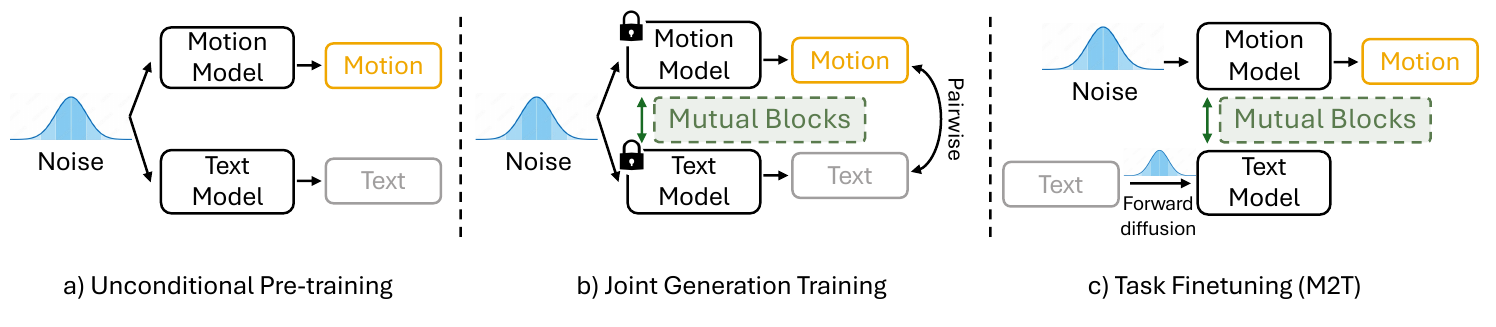}
\caption{Training stages of the PackDiT model, illustrating the various phases, including a) unconditional pre-training, b) joint generation training, and c) task fine-tuning.}
\label{fig:training}
\end{figure*}

\begin{figure*}[t]
\setcounter{figure}{0}
\renewcommand\figurename{Algorithm}
\caption{The \textbf{pseudo-code} of different training stages of PackDiT depends on different tasks, \eg, unconditional pre-train, Text-to-Motion, and Motion-to-Text.}
\vspace{-5mm}
\begin{minipage}[b]{0.49\textwidth}
\begin{algorithm}[H]
    \footnotesize
    \centering
    \caption{\small Unconditional Pre-train}
    \begin{algorithmic}
        \Require Motion tokens, $\mathcal{M}$, Text tokens, $\mathcal{T}$
        \State \hspace{-1em}\textbf{Repeat}
        \State $t_\mathcal{M}, t_\mathcal{T} \sim \text{Uniform}(1,\dots,t)$
        \State $\epsilon_{\mathcal{M}}, \epsilon_{\mathcal{T}} \sim \mathcal{N}(0, 1)$
        \State $\mathcal{M}_{t_\mathcal{M}} = \sqrt{\overline{\alpha}_{t_\mathcal{M}}}\mathcal{M} + \sqrt{1 - \overline{\alpha}_{t_\mathcal{M}}}\epsilon_{\mathcal{M}}$
        \State $\mathcal{T}_{t_\mathcal{T}} = \sqrt{\overline{\alpha}_{t_\mathcal{T}}}\mathcal{T} + \sqrt{1 - \overline{\alpha}_{t_\mathcal{T}}}\epsilon_{\mathcal{T}}$
        \State Take gradient step on
        \State \hspace{1em}$\nabla_{D_\mathcal{M}}\|\epsilon - D_\mathcal{M}(\mathcal{M}_{t_\mathcal{M}}, t_\mathcal{M})\|^2_2$
        \State \hspace{1em}$\nabla_{D_\mathcal{T}}\|\epsilon - D_\mathcal{T}(\mathcal{T}_{t_\mathcal{T}}, t_\mathcal{T})\|^2_2$
        \State \hspace{-1em}\textbf{until} converge
    \end{algorithmic}
\end{algorithm}
\end{minipage}
\hfill
\begin{minipage}[b]{0.49\textwidth}
\begin{algorithm}[H]
    \footnotesize
    \centering
    \caption{\small Joint Generation Training}
    \begin{algorithmic}
        \Require Paired Motion tokens, $\mathcal{M}$, Text tokens, $\mathcal{T}$
        \State \hspace{-1em}\textbf{Repeat}
        \State $t_\mathcal{M} = t_\mathcal{T} \sim \text{Uniform}(1,\dots,t)$
        \State $\epsilon_{\mathcal{M}}, \epsilon_{\mathcal{T}} \sim \mathcal{N}(0, 1)$
        \State $\mathcal{M}_{t_\mathcal{M}} = \sqrt{\overline{\alpha}_{t_\mathcal{M}}}\mathcal{M} + \sqrt{1 - \overline{\alpha}_{t_\mathcal{M}}}\epsilon_{\mathcal{M}}$
        \State $\mathcal{T}_{t_\mathcal{T}} = \sqrt{\overline{\alpha}_{t_\mathcal{T}}}\mathcal{T} + \sqrt{1 - \overline{\alpha}_{t_\mathcal{T}}}\epsilon_{\mathcal{T}}$
        \State Take gradient step on
        \State \hspace{1em}$\nabla_{D_\mathcal{M}}\|\epsilon - D_\mathcal{M}(\mathcal{M}_{t_\mathcal{M}}, t_\mathcal{M}), D_\mathcal{T}(\mathcal{T}_{t_\mathcal{T}}, t_\mathcal{T}))\|^2_2$
        \State \hspace{1em}$\nabla_{D_\mathcal{T}}\|\epsilon - D_\mathcal{T}(\mathcal{T}_{t_\mathcal{T}}, t_\mathcal{T}, D_\mathcal{M}(\mathcal{M}_{t_\mathcal{M}}, t_\mathcal{M}))\|^2_2$
        \State \hspace{-1em}\textbf{until} converge
    \end{algorithmic}
\end{algorithm}
\end{minipage}


\renewcommand\figurename{Algorithm}
\begin{minipage}[b]{0.49\textwidth}
\begin{algorithm}[H]
    \footnotesize
    \centering
    \caption{\small Text-to-Motion Training}
    \begin{algorithmic}
        \vspace{0.2mm}
        \Require Paired Motion tokens, $\mathcal{M}$, Text tokens, $\mathcal{T}$
        \State \hspace{-1em}\textbf{Repeat}
        \State $t_\mathcal{M} \sim \text{Uniform}(1,\dots,t)$
        \State $\epsilon_{\mathcal{M}} \sim \mathcal{N}(0, 1)$
        \State $\mathcal{M}_{t_\mathcal{M}} = \sqrt{\overline{\alpha}_{t_\mathcal{M}}}\mathcal{M} + \sqrt{1 - \overline{\alpha}_{t_\mathcal{M}}}\epsilon_{\mathcal{M}}$
        \State Take gradient step on
        \State \hspace{1em}$\nabla_{D_\mathcal{M}}\|\epsilon - D_\mathcal{M}(\mathcal{M}_{t_\mathcal{M}}, t_\mathcal{M}, D_\mathcal{T}(\mathcal{T}, 0)\|^2_2$
        \State \hspace{-1em}\textbf{until} converge
    \end{algorithmic}
\end{algorithm}
\end{minipage}
\hfill
\begin{minipage}[b]{0.49\textwidth}
\begin{algorithm}[H]
    \footnotesize
    \centering
    \caption{\small Joint Training}
    \begin{algorithmic}
        \vspace{1.8mm}
        \Require Motion tokens, $\mathcal{M}$, Text tokens, $\mathcal{T}$
        \State \hspace{-1em}\textbf{Repeat}
        \State task $= \text{RandomChoise}(\text{t2m, m2t, uncond, joint})$
        \State $t_\mathcal{M}, t_\mathcal{T} \sim \text{Uniform}(1,\dots,t)$
        \State $\epsilon_{\mathcal{M}}, \epsilon_{\mathcal{T}} \sim \mathcal{N}(0, 1)$
        \State Train$(D_\mathcal{M}, D_\mathcal{T}, \mathcal{M}, \mathcal{T}, t_\mathcal{M}, t_\mathcal{T}, \epsilon_{\mathcal{M}}, \epsilon_{\mathcal{T}}, \text{task})$
        \State \hspace{-1em}\textbf{until} converge
        \vspace{1.9mm}
    \end{algorithmic}
\end{algorithm}
\end{minipage}
\label{alg:training}
\end{figure*}

\subsection{Training Recipe}
\noindent\textbf{Step 1: Unconditional Pre-train.} Since there are two independent DiTs in PackDiT, we can apply unconditional Pre-train on both of them to get better initial weights for the subsequent tasks. As shown in Algorithm~\ref{alg:training}, during unconditional pre-train, motion tokens and text tokens are sampled from the training dataset and are fed to $D_\mathcal{M}$ and $D_\mathcal{T}$ separately for standard unconditional diffusion training.

\vspace{3pt}
\noindent\textbf{Step 2: Joint Generation Training.} To make PackDiT able to conduct joint Motion and Text generation and better align the features from two modalities, we conduct joint generation training. We sample the same timestep $t_\mathcal{M} = t_\mathcal{T} \sim \text{Uniform}(1,\dots,t)$ and the mutual blocks are enabled during training. Thus, both DiTs are trained together for feature alignment and joint generation. 

\vspace{3pt}
\noindent\textbf{Step 3: Motion-to-Text and Text-to-Motion Fine-tuning.} As demonstrated in Algorithm~\ref{alg:training}, during training for either the Motion-to-Text or Text-to-Motion task, only the generating modality undergoes forward diffusion. In contrast, the condition modality is directly fed to the conditional DiT after Patchify, with the timestep set to $0$. Consequently, through the mutual blocks between the two DiTs, PackDiT effectively performs reliable conditional generation.

\begin{table*}[t]
\caption{Comparison of three motion-related tasks on HumanML3D~\cite{guo2022humanml3d} dataset. The evaluation metrics are computed using the encoders introduced in \cite{guo2022humanml3d}. $^\dagger$ indicates that LMM~\cite{zhang2024large} is trained with additional data.}
\resizebox{\textwidth}{!}{%
\begin{tabular}{lcccccccc}
\toprule
\multirow{2}{*}{Methods} & \multirow{2}{*}{Source} &
\multicolumn{3}{c}{Text-to-Motion}& 
\multicolumn{2}{c}{Motion Prediction}&
\multicolumn{2}{c}{Motion In-between}
\\ \cmidrule(lr){3-5} \cmidrule(lr){6-7} \cmidrule(lr){8-9} 
& &
R@1~($\uparrow$) 
 & FID~($\downarrow$)
&
DIV~($\uparrow$) 
& FID~($\downarrow$) & DIV~($\uparrow$)  & FID~($\downarrow$) & DIV~($\uparrow$)
\\ \midrule
Real Data & - &
 $0.511^{\pm.003}$ &
  $0.002^{\pm.000}$ &
  $9.503^{\pm.065}$ 
  & 0.002 & 9.503 & 0.002 & 9.503
  \\ \midrule
MLD \cite{chen2023mld} & CVPR'23 &
 $0.481^{\pm.003}$ &
  ${0.473}^{\pm.013}$ &

  $\underline{9.724}^{\pm.082}$
  & \multicolumn{1}{c}{-}& \multicolumn{1}{c}{-}& \multicolumn{1}{c}{-} &
  \multicolumn{1}{c}{-}
  \\
T2M-GPT \cite{zhang2023t2mgpt}& CVPR'23 &
   ${0.491}^{\pm.003}$&
  ${0.116}^{\pm.004}$ &
  $\textbf{9.761}^{\pm.081}$ 
  & \multicolumn{1}{c}{-}& \multicolumn{1}{c}{-} & \multicolumn{1}{c}{-}& \multicolumn{1}{c}{-}
  \\
TM2T \cite{guo2022tm2t} & ECCV'22 &
 $0.424^{\pm.017}$ &
  ${1.501^{\pm.003}}$ &
  $8.589^{\pm.076}$
  & \multicolumn{1}{c}{-}& \multicolumn{1}{c}{-}& \multicolumn{1}{c}{-}& \multicolumn{1}{c}{-}
  \\
MDM \cite{tevet2023mdm}& ICLR'23 &
 $0.320^{\pm005}$ &
${0.544}^{\pm.044}$ &
  ${9.559}^{\pm.086}$ &
  $6.031$ & $7.813$& $2.698 $&$ 8.420$
  \\
MotionGPT \cite{jiang2024motiongpt} & NeurIPS'23 & 
${0.492}^{\pm.003}$ &
${0.232}^{\pm.008}$ & 
${9.528}^{\pm.071}$ &
${0.905}$ &${8.972}$  & $0.214$&$\textbf{9.560}$
   \\ 
LMM-Tiny$^\dagger$ \cite{zhang2024large}& ECCV'24 &
$0.496^{\pm.002}$ &
$0.415^{\pm.002}$ & 
$9.176^{\pm.074}$ &
\multicolumn{1}{c}{-}
&\multicolumn{1}{c}{-}  & \multicolumn{1}{c}{-}&\multicolumn{1}{c}{-} \\
LMM-Small$^\dagger$ \cite{zhang2024large}& ECCV'24 &
$\underline{0.505}^{\pm.002}$ &
$\underline{0.227}^{\pm.002}$ & 
$9.295^{\pm.076}$ &
\multicolumn{1}{c}{-}
&\multicolumn{1}{c}{-}  & \multicolumn{1}{c}{-}&\multicolumn{1}{c}{-} \\
TMT \cite{qian2024tmt}& ECCV'24 &
$ 0.464^{\pm \text{ unk.}}$ &
$ 0.310^{\pm \text{ unk.}}$ & 
$ 9.191^{\pm \text{ unk.}}$ &
\multicolumn{1}{c}{-} &\multicolumn{1}{c}{-}  & \multicolumn{1}{c}{-}&\multicolumn{1}{c}{-} \\
\midrule
PackDiT-Tiny & Ours &
$0.498^{\pm.003}$ &
$0.232^{\pm.006}$ & 
$9.381^{\pm.071}$ &
$\underline{0.764}$ &$\textbf{9.140}$  & $\underline{0.131}$&$8.974$ \\
PackDiT-Small & Ours &
$\boldsymbol{0.510}^{\pm.003}$ &
$\boldsymbol{0.106}^{\pm.006}$ & 
$9.680^{\pm.078}$ &
$\boldsymbol{0.701}$ & $\underline{9.046}$  & $\boldsymbol{0.119}$ & $\underline{9.114}$ \\
   \bottomrule
\end{tabular}%
}
\label{tab:uniform}
\end{table*}

\vspace{3pt}
\noindent\textbf{Step 4: Joint Fine-Tuning.} To train PackDiT on all four tasks, we employ a joint training approach, assigning a certain probability to each task at each iteration. For optimal performance in the evaluations detailed in Section~\ref{sec:exp}, we further fine-tune the model on specific tasks, \ie, Text-to-Motion and Motion-to-Text generation, after the initial joint training phase. 
Therefore, the final training objective of PackDiT is to minimize the loss $\mathcal{L}$:
\begin{equation}
    \mathcal{L} = \mathcal{L}_\mathcal{M} + \lambda \cdot \mathcal{L}_\mathcal{T},
\end{equation}

\noindent where $\lambda$ is the term to balance the two objectives.

\section{Experimental Results}
\label{sec:exp}

\begin{table}
\centering
\caption{Performance comparison on the \textbf{Motion-to-Text} task for the HumanML3D~\cite{guo2022humanml3d} dataset. All evaluation metrics are computed using encoders from \cite{guo2022humanml3d}.}
\resizebox{0.98\linewidth}{!}{
\begin{tabular}{l c c c}
\toprule
Methods & R@3~($\uparrow$) & BLEU@4~($\uparrow$) & CIDEr~($\uparrow)$   \\
\midrule
Real Data & 0.828 & - & - \\
\midrule
\textit{LLM-based} \vspace{3pt} \\
TM2T~\cite{guo2022tm2t} & $0.823$ & ${7.00}$ & $16.8$ \\
MotionGPT~\cite{jiang2024motiongpt} & $\boldsymbol{0.827}$ & $\boldsymbol{12.47}$ & $\boldsymbol{29.2}$ \\
\midrule
\textit{Diffusion-based} \vspace{3pt} \\
PackDiT-Tiny & $0.810$ & $6.86$ & $13.6$ \\
PackDiT-Small & $\underline{0.825}$ & $\underline{11.82}$ & $\underline{25.5}$\\
\bottomrule
\end{tabular}
}
\end{table}

\subsection{Evaluation Metrics and Datasets}
\vspace{-6pt}
\paragraph{Evaluation Metrics.} Following~\cite{guo2022humanml3d, tevet2023mdm, guo2022tm2t, jiang2024motiongpt}, Frechet Inception Distance (FID) is our primary metric for motion quality evaluation, which evaluates the feature distribution similarity between generated and real motions as detailed in \cite{guo2022humanml3d}. Meanwhile, to measure the diversity of the generated motions, we use the Diversity (DIV) metric, which calculates the variance in features extracted from the motions as used in \cite{guo2022humanml3d}. For text-motion retrieval evaluation, the accuracy of matching motions to their corresponding textual descriptions is assessed using the motion-retrieval precision (R Precision) metric, based on the feature space from \cite{guo2022humanml3d}, and measured by Top 1/2/3 retrieval accuracy. To evaluate the quality of generated motion captions, we adopt linguistic metrics from natural language processing studies as outlined in \cite{guo2022tm2t}, including BLEU \cite{papineni2002bleu}, and CIDEr \cite{vedantam2015cider}.

\paragraph{Dataset.} The HumanML3D dataset~\cite{guo2022humanml3d} is a comprehensive repository of 3D human motion sequences curated to advance research in human motion analysis and generation. It encompasses a diverse range of activities—including walking, running, dancing, and more complex actions—sourced from the AMASS dataset~\cite{mahmood2019amass}. Comprising 14,616 motion sequences of durations between 2 and 10 seconds, each sequence is accompanied by multiple detailed textual annotations, enhancing its applicability for tasks such as text-to-motion and motion-to-text generation. The dataset incorporates a variety of actors to ensure broad representation across human movement patterns.



\begin{table*}[t]
\centering
\caption{\textbf{Ablation study} on PackDiT. Evaluated PackDiT-Tiny on the HumanML3D dataset with the Text-to-Motion task. $-$ means the hyperparameter or setup does not change compared with the baseline.}
\resizebox{0.7\linewidth}{!}{
\begin{tabular}{l|cccc|cc}
\toprule
\multirow{2}{*}[-1pt]{Experiments}& \multicolumn{4}{c|}{Ablation Settings} & \multicolumn{2}{c}{Text-to-Motion} \\
 & Dim$_{P}$ & Text Encoder & Patch Size & w. Uncond~ & ~FID~($\downarrow$)  & R@1~($\uparrow$)~\\ 
\midrule
Baseline & 64 & BERT~\cite{devlin2018bert} & 1 & $\times$ & $0.274$ & $0.493$ \\
\midrule
\multirow{2}{*}[-1pt]{Projection Dim.}& $128$ & BERT~\cite{devlin2018bert} & 1 & $\times$ & $0.264$ & $0.493$ \\
& $256$ & BERT~\cite{devlin2018bert} & 1 & $\times$ & ${0.251}$ & ${0.495}$ \\
\midrule
Text Encoder& $64$ & T5~\cite{raffel2020t5} & 1 & $\times$ & $0.618$ & $0.472$ \\
\midrule
\multirow{2}{*}[-1pt]{Patch Size}& $64$ & BERT~\cite{devlin2018bert} & 2 & $\times$ & $0.571$ & $0.481$ \\
& $64$ & BERT~\cite{devlin2018bert} & 4 & $\times$ & $1.483$ & - \\
\midrule
PackDiT-Tiny & $64$ & BERT~\cite{devlin2018bert} & 1 & $\checkmark$ & $\boldsymbol{0.232}$  & $\boldsymbol{0.498}$ \\
\bottomrule
\end{tabular}
}
\label{tab:ablation}
\end{table*}

\subsection{Training and Evaluation Details}

We utilize a single NVIDIA A100 to train and evaluate PackDiT, which is developed on OpenDiT~\cite{zhao2024opendit} and MotionGPT~\cite{jiang2024motiongpt}. The number of parameters for each DiT of proposed PackDiT Tiny and Small is around $75$M and $120$M, which are similar to the setup of LMM~\cite{zhang2024large}. With the batch size as $128$, PackDiT is trained with the Adam~\cite{kingma2014adam} optimizer, and the learning rate is set to $10^{-4}$. The patch size is set to $1$ for both $D_\mathcal{M}$ and $D_\mathcal{T}$ during evaluation, and more patch size setup is discussed in Ablation Study~\ref{subsec:abl}. As mentioned in Sec~\ref{sec:method}, a text encoder, a text decoder, and a projection model are used in the Text Generation pipeline. Following~\cite{bao2023unidiffuser}, BERT~\cite{devlin2018bert}, and GPT-2~\cite{radford2019gpt2} are used as the text encoder and text decoder, respectively. The projection model, $P$, is trained with projection dimension $64$ when the encoder and decoder models are frozen. The unconditional pre-train takes around $10$ epochs. Then, the PackDiT is jointly trained with all tasks for $200$ epochs.
To achieve the best performance of PackDiT, the models used for evaluation are fine-tuned on specific tasks for $300$ epochs after joint training with all tasks. The motion representations of PackDiT follows~\cite{guo2022humanml3d, jiang2024motiongpt} for fair comparison.

\subsection{Experimental Results of All Tasks}

\textbf{Evaluation on Motion-related Tasks.}
Our proposed method is compared with other SOTA methods on the HumanML3D \cite{guo2022humanml3d} dataset. As shown in Table \ref{tab:uniform}, our PackDiT-Tiny and PackDiT-Small achieve $0.498$ and $0.504$ R@1 on text-to-motion task, which demonstrates a comparable performance with previous SOTA method LMM \cite{zhang2024large}, while using a smaller amount of training data. In other tasks like motion prediction and motion in-between tasks, PackDiT achieves $0.701$ and $0.119$ FID scores, outperforming MotionGPT \cite{jiang2024motiongpt} by a large margin.

\paragraph{Evaluation on Motion-to-Text.}
We also evaluate PackDiT's performance on the motion-to-text task. Despite utilizing a diffusion-based backbone, our proposed method demonstrates effective motion-to-text generation capabilities. PackDiT-small achieves an R@3 of $0.784$, a BLEU@4 score of $8.12$, and a CIDEr score of $15.5$. These results are comparable to those of LLM-based methods, highlighting PackDiT's competitive performance despite the inherent challenges of using a diffusion model for language tasks.

\paragraph{Visualization Results.} As shown in Fig~\ref{fig:vis}, we present additional Text-to-Motion generation results. These results demonstrate that our method, PackDiT, is capable of generating diverse and reliable motion sequences. The generated motions exhibit a wide variety of actions and behaviors, accurately reflecting the nuances of the input text descriptions. The visualization videos are attached to the appendix.

\subsection{Ablation Studies}
\label{subsec:abl}

\begin{figure*}[t]
    \centering
    \setcounter{figure}{2}
    \includegraphics[width=0.98\linewidth]{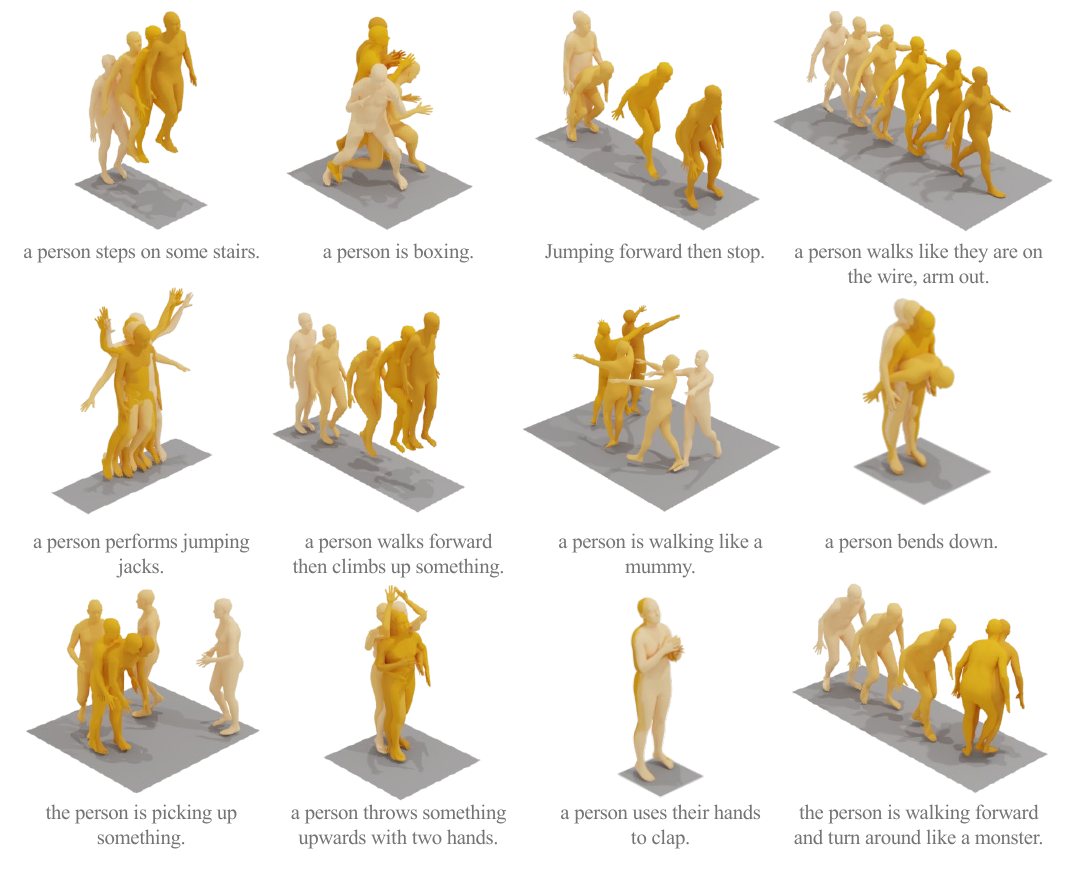}
    \caption{Visualization results of Text-to-Motion Generation via PackDiT.}
    \label{fig:vis}
\end{figure*}

To further analyze the PackDiT, we conduct ablation study on several hyperparameters, training strategies and alternative models used by us. All ablation studies are evaluated on Text-to-Motion tasks with PackDiT-Tiny as default.

\paragraph{Target Dimension of Projection Model.} As shown in Tab~\ref{tab:ablation}, the best performance is achieved when the target dimension of the projection model, $P$, is set to $256$. $128$ and $256$ are closer to the dimension of BERT's hidden states and save more information from the original text tokens. However, based on the trade-off between accuracy and efficiency, we choose Dim$_P = 64$ when we are training PackDiT-Tiny. 

\paragraph{Text Encoder.} Following~\cite{bao2023unidiffuser}, we utilize BERT~\cite{devlin2018bert} as our text encoder for the text generation pipeline. To further investigate the PackDiT, T5~\cite{raffel2020t5}, an Encoder-Decoder based Large Language Model, is applied to our text pipeline for a comparison. Only the encoder part of the T5-base is integrated. As illustrated in Tab~\ref{tab:ablation}, the performance of Text-to-Motion with BERT surpasses the T5 version by a remarkable margin since the pre-train weights of T5 are based on translation, summarization, question answering, and \etc, which may not be suitable for motion-related tasks.

\paragraph{Patch Size.} We change the patch size of motion diffusion model and conduct an ablation study. According to Tab~\ref{tab:ablation}, the patch size $1$ provides the best performance while patch size $4$ significantly impacts the performance. We find that once the dimension of input tokens is similar to the hidden dimension of DiTs, the generation results are not reliable.

\paragraph{Unconditional Pre-train.} As indicated in Table~\ref{tab:ablation}, the Unconditional Pre-train improves the final Text-to-Motion Generation performance. This pre-training phase allows PackDiT to develop a better understanding of the underlying data distribution, which is crucial for generating realistic and coherent motions. By initializing the model with weights that are already attuned to the data characteristics, the subsequent training process is more efficient and effective. We plan to utilize more unpaired motion sequences and motion descriptions for future works and further improve the effectiveness of the Unconditional Pre-train.

\begin{table}
\centering
\caption{The comparison of the number of parameters. The PackDiT includes both text and motion diffusion models.}
\resizebox{0.6\linewidth}{!}{
\begin{tabular}{l c c}
\toprule
Methods & Arch. & \# Param. \\
\midrule
MotionGPT~\cite{jiang2024motiongpt} & AR & $248$M \\
LMM-Tiny~\cite{zhang2024large} & Diffusion & $90$M\\
LMM-Small~\cite{zhang2024large} & Diffusion & $160$M\\
\midrule
PackDiT-Tiny & Diffusion & $72$M\\
PackDiT-Small & Diffusion & $229$M\\
\bottomrule
\end{tabular}
}
\label{tab:num_param}
\end{table}
\paragraph{Number of Parameters}
To make a fair comparison with other SOTA motion generation models, we compare the number of parameters of PackDiT with other methods. As shown in Tab~\ref{tab:num_param}, we compare the number of parameters with MotionGPT~\cite{jiang2024motiongpt} and LMM~\cite{zhang2024large}, which shows a fair comparison with other SOTA methods.

\section{Limitations}
 
The performance of PackDiT heavily relies on the quality and diversity of the HumanML3D dataset. Limited data variety may hinder the generalization of the model. Also, current evaluation metrics such as FID and Recall may not fully capture the quality and realism of generated motions and texts, suggesting a need for more comprehensive performance assessment methods.



\section{Conclusion}

In this work, we presented \textbf{PackDiT}, a novel diffusion-based framework for joint human motion and text generation. PackDiT’s unique dual Diffusion Transformer (DiT) architecture with mutual blocks enables efficient handling of multiple generation tasks, including text-to-motion, motion-to-text, motion prediction, and motion in-between generation within a unified model structure. This approach addresses critical limitations in previous models, which often restricted generation to single modalities or required complex configurations.
Extensive experiments on the HumanML3D dataset demonstrate that PackDiT achieves state-of-the-art results, particularly in text-to-motion generation with an FID score of $0.106$, as well as strong performance in motion prediction and in-between tasks. The mutual prompting mechanism facilitates enhanced information exchange, enabling PackDiT to produce high-quality, diverse outputs that closely align with human-generated data.
PackDiT thus establishes a flexible and robust foundation for multi-modal generation, with broad implications for synthetic data creation, immersive environments, and human-computer interaction applications, setting a new standard for joint motion and text synthesis.


{
    \small
    \bibliographystyle{ieeenat_fullname}
    \bibliography{main}
}

\clearpage
\renewcommand{\thesection}{\Alph{section}}
\renewcommand\thefigure{\Alph{section}\arabic{figure}}
\renewcommand\thetable{\Alph{section}\arabic{table}}
\setcounter{page}{1}
\setcounter{section}{0}
\setcounter{figure}{0}
\setcounter{table}{0}

\twocolumn[
{
        \centering
        \Large
        \textbf{\thetitle}\\
        \vspace{0.5em}Supplementary Material \\
        \vspace{1.0em}
}
{
    \renewcommand\twocolumn[1][]{#1}
    \begin{center}
    \centering
    \renewcommand\thefigure{A1}
    \includegraphics[width=\linewidth]{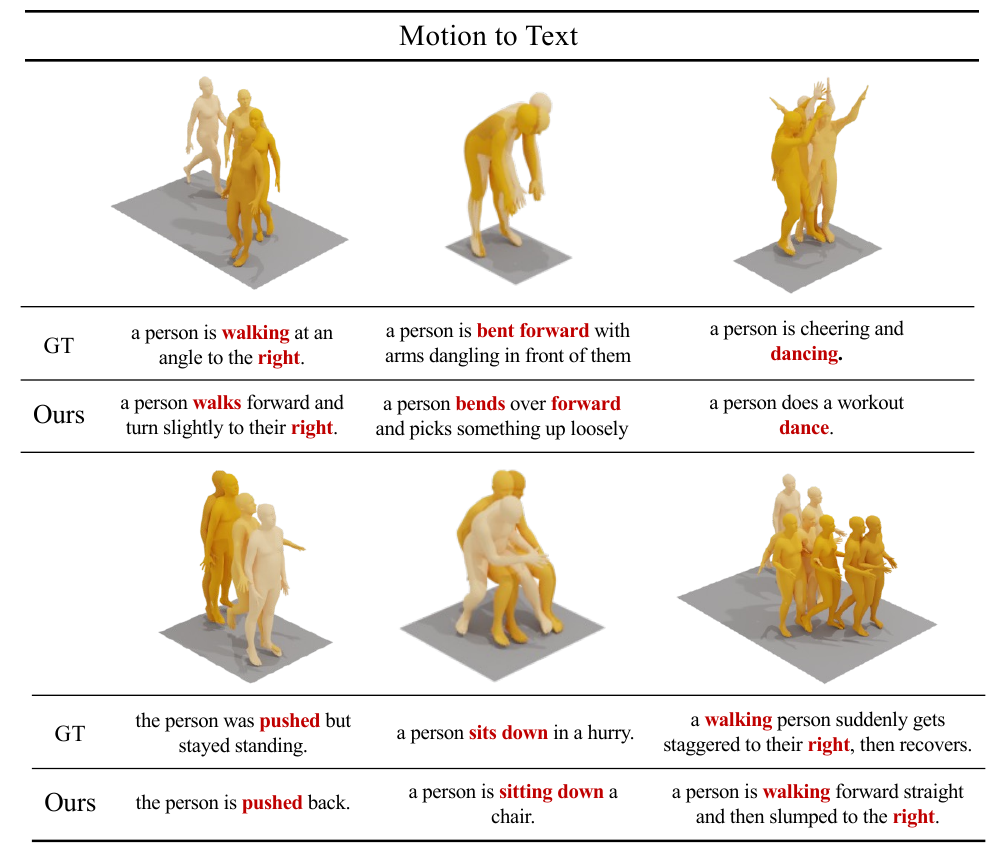}
    \captionof{figure}{More Motion-to-Text visualization results of PackDiT on HumanML3D dataset.}
    \vspace{12pt}
    \label{fig:m2t_supp_vis}
\end{center}
}]

\section{More Qualitative Results}
\begin{figure*}
    \centering
    \includegraphics[width=0.9\linewidth]{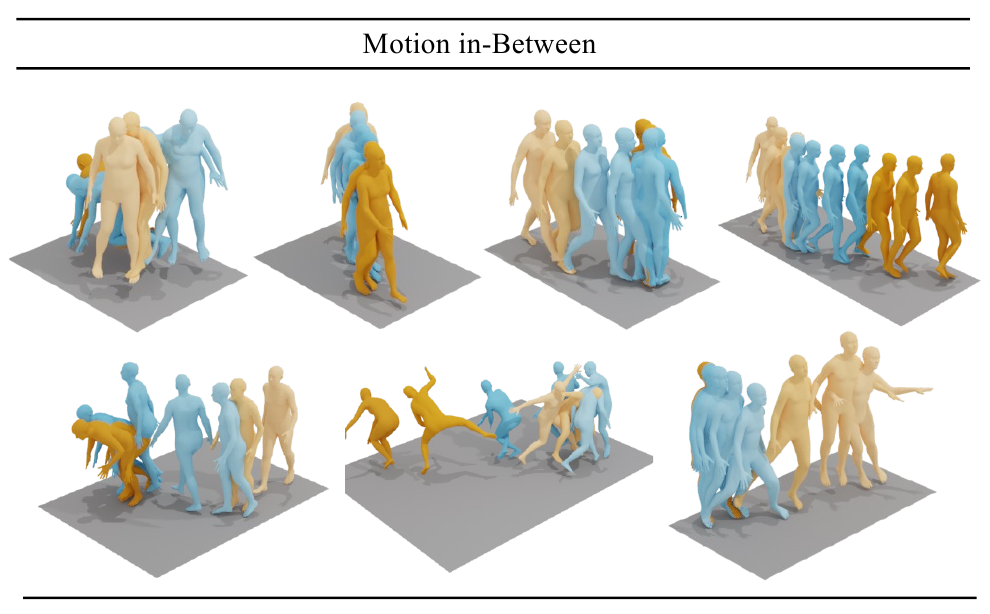}
    \caption{More Motion in-Between visualization results of PackDiT. The orange avatars are from the ground truth motion, while the blue ones are generated by PackDiT.}
    \label{fig:completion_supp_vis}
\end{figure*}

\begin{figure*}
    \centering
    \includegraphics[width=0.9\linewidth]{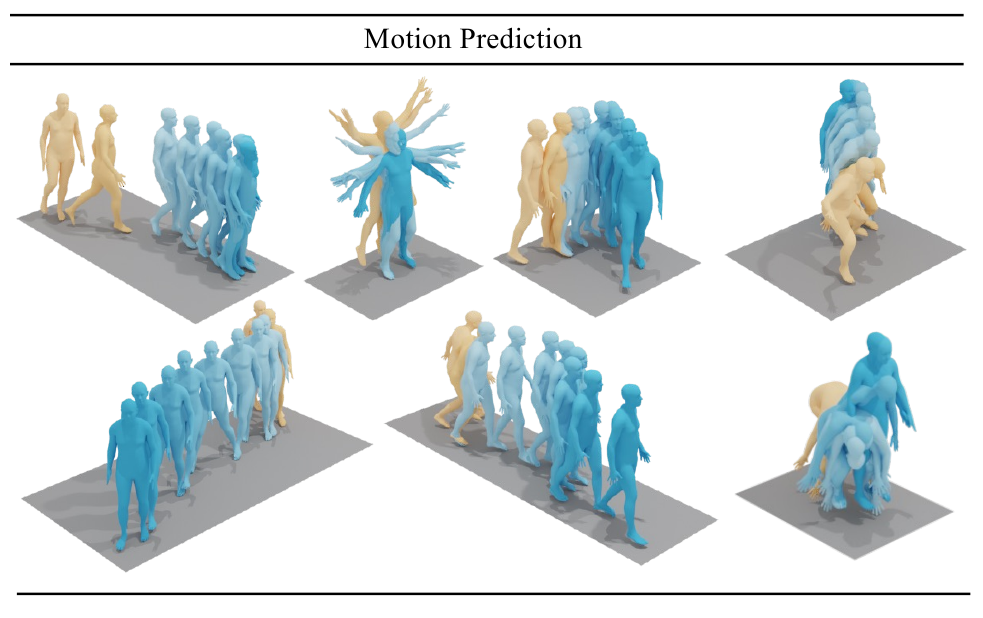}
    \caption{More Motion Prediction visualization results of PackDiT. The orange avatars are from the ground truth motion, while the blue ones are generated by PackDiT.}
    \label{fig:pred_supp_vis}
\end{figure*}

\begin{figure*}
    \centering
    \includegraphics[width=\linewidth]{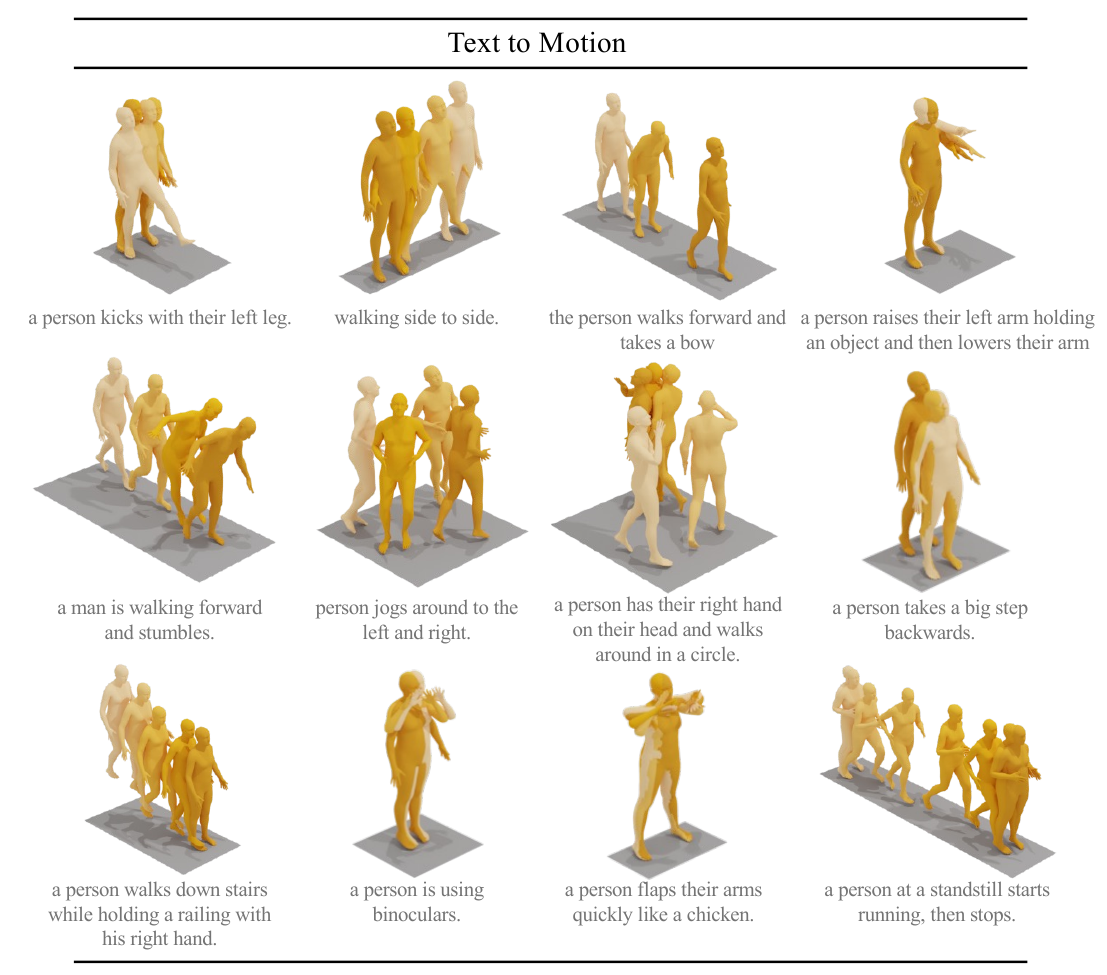}
    \caption{More Text-to-Motion visualization results of PackDiT.}
    \label{fig:supp_vis}
\end{figure*}

Motion-to-Text visualization results are presented in \cref{fig:m2t_supp_vis}, demonstrating the effectiveness of our proposed method. Similarly, extended Text-to-Motion visualization results are shown in \cref{fig:supp_vis}, highlighting the ability of PackDiT to generate diverse and temporally stable motions that adhere closely to the given descriptions. Visualization results for the Motion In-Between task are provided in \cref{fig:completion_supp_vis}, emphasizing PackDiT's capability to produce smooth and contextually coherent intermediate motions. Moreover, \cref{fig:pred_supp_vis} showcases the results of the Motion Prediction task, illustrating the model’s ability to accurately predict plausible future motions based on prior sequences. These results demonstrate that PackDiT achieves high-quality performance across four tasks, including Motion-to-Text, Text-to-Motion, Motion In-Between, and Motion Prediction, underscoring its stability and robustness.

\end{document}